\documentclass{article}
\usepackage{microtype}
\usepackage{graphicx}
\usepackage{subcaption}
\usepackage{booktabs} 
\usepackage{multirow} 

\usepackage{hyperref}

\usepackage[preprint]{icml2026}

\usepackage[table]{xcolor}
\usepackage{pifont}    
\newcommand{\cmark}{\ding{51}} 
\newcommand{\xmark}{\ding{55}} 

\usepackage{amsmath}
\usepackage{amssymb}
\usepackage{mathtools}
\usepackage{amsthm}
\usepackage{enumitem}

\usepackage{tcolorbox}

\usepackage[capitalize,noabbrev]{cleveref}
\crefname{appendix}{Appendix}{Appendices}
\Crefname{appendix}{Appendix}{Appendices}
\theoremstyle{plain}

\theoremstyle{definition}

\theoremstyle{remark}

\usepackage[textsize=tiny]{todonotes}

\usepackage{diagbox}

\icmltitlerunning{Exploring Pass-Rate Reward in Reinforcement Learning for Code Generation}

\begin{document}

\twocolumn[
\icmltitle{Exploring Pass-Rate Reward in Reinforcement Learning for Code Generation}



  \icmlsetsymbol{equal}{*}

  \begin{icmlauthorlist}
    \icmlauthor{Xin-Ye Li}{equal,nju,ai}
    \icmlauthor{Ren-Biao Liu}{equal,nju,ai}
    \icmlauthor{Yun-Ji Zhang}{nju,ai}
    \icmlauthor{Hui Sun}{nju,ai}
    \icmlauthor{Zheng Xie}{nju}
    \icmlauthor{Ming Li}{nju,ai}
  \end{icmlauthorlist}

  \icmlaffiliation{nju}{State Key Laboratory of Novel Software Technology, Nanjing University, Nanjing 210023, China}
  \icmlaffiliation{ai}{School of Artificial Intelligence, Nanjing University, Nanjing 210023, China}

  \icmlcorrespondingauthor{Ming Li}{lim@lamda.nju.edu.cn}

  \icmlkeywords{Reinforcement Learning, Code Generation, Large Language Models}

  \vskip 0.3in
]



\printAffiliationsAndNotice{\icmlEqualContribution}

\begin{abstract}
Reinforcement learning (RL) from unit-test feedback has become a standard post-training recipe for improving large language models (LLMs) on code generation.
However, the \emph{pass-all-tests} binary reward can be sparse, yielding no learning signal on challenging problems where none of the sampled solutions passes all tests.
A common remedy is to use the \emph{test-case pass rate} as a surrogate reward.
In this work, we study pass-rate rewards in \emph{critic-free} RL for code generation (e.g., GRPO and RLOO) and report a consistent pattern across base models and algorithms: despite alleviating reward sparsity, pass-rate rewards do not reliably improve final performance over binary rewards in rigorous controlled experiments.
To understand this discrepancy, we analyze reward density and the resulting gradient directions.
We find that pass-rate rewards are denser, but the induced gradient updates do not consistently move probability mass toward full-pass solutions.
This arises because test-case pass rate is a miscalibrated surrogate for progress toward full correctness, and partial-pass solutions within the same group can induce conflicting gradient directions that cancel out.
Overall, our results suggest that, in critic-free RL, pass-rate rewards are insufficient to improve code generation and motivate reward designs that better align optimization with the goal of full correctness.
\end{abstract}

\section{Introduction}\label{sec:introduction}

Recent advancements in large language models~(LLMs) have revolutionized code generation, significantly enhancing coding productivity. 
Furthermore, reinforcement learning with verifiable rewards~(RLVR), a prominent post-training paradigm for enhancing LLM reasoning and coding capabilities, has attracted considerable research attention~\cite{jaech2024openai-o1,guo2025deepseek-r1,team2025kimi}.

In software development, unit testing provides a standard methodology for validating code correctness.
Correspondingly, code generation tasks typically employ a binary signal to assess correctness: whether the generated code passes all test cases.
Consequently, most RL for code generation methods naturally leverage this signal as a binary reward~\cite{li2022competition,liu2023rltf}.
However, such binary reward may become prohibitively sparse on challenging problems, where none of the generated programs pass all test cases in a rollout group, thereby providing no effective learning signal and reducing training efficiency~\cite{yu2025dapo,hong2025glm4}.
Therefore, several recent works~\cite{xiaomi2025mimo,cui2025process,sun2025deltas} attempt to adopt the test-case pass rate as a surrogate reward to alleviate the sparsity of binary rewards, providing non-zero learning signals for partial-pass solutions.
However, the comparative effects of binary rewards \emph{vs.} pass-rate rewards in RL for code generation remain largely unexplored.
While pass-rate rewards intuitively provide denser learning signals that should facilitate optimization, their actual impact on final performance and learning dynamics is unclear.

In this work, we conduct rigorous controlled experiments across different pretrained base LLMs, two representative \emph{critic-free} RL algorithms (GRPO and RLOO), and multiple reward designs, including binary rewards, pass-rate rewards, and variants such as test-case reweighted pass-rate rewards and two-stage combinations.
Across extensive experiments, we uncover a counterintuitive result: \textbf{in this critic-free RL regime, pass-rate rewards do not achieve higher performance ceilings than binary rewards}, despite providing denser learning signals that should facilitate more effective optimization.

To shed light on this counterintuitive result, we examine both the \emph{density} and the \emph{direction} of the learning signal provided by pass-rate rewards.
We first confirm that pass-rate rewards are dense, with many rollouts receiving intermediate scores rather than collapsing to near-binary feedback.
However, denser feedback does not necessarily mean better guidance: we find that \textbf{test-case pass rate is not a reliable surrogate for proximity to full correctness.}
In particular, pass-rate rewards can reinforce solutions that exploit shortcuts (e.g., brute-force heuristics or overfitting to the provided test suite) to achieve high pass rates, while penalizing near-correct solutions that require minor fixes to pass all tests.
Under this miscalibration, the policy-gradient updates induced by partial-pass solutions are often weak and noisy. When no rollout fully passes all tests, updates point in different directions and tend to cancel out at the group level, yielding little overall progress toward full correctness.

Our main contributions are summarized as follows:
\begin{itemize}[leftmargin=*,noitemsep,topsep=0pt]
    \item We conduct controlled experiments across multiple base LLMs, two critic-free RL algorithms (GRPO and RLOO), and reward designs, and find that, in this regime, pass-rate rewards do not consistently improve final performance over binary rewards.
    \item We provide a mechanistic analysis of learning dynamics under pass-rate rewards in critic-free RL, identifying a miscalibration between test-case pass rate and true progress toward full correctness and showing how this miscalibration can lead to weak optimization signals toward correct solutions.
    \item We distill practical implications for critic-free RL on code generation, clarifying when pass-rate rewards can be helpful yet misleading, and motivating reward designs that better align optimization with the goal of full correctness.
\end{itemize}

The remainder of this paper is organized as follows: \Cref{sec:preliminary} introduces preliminaries on RL from unit-test feedback and relevant policy-gradient methods; \Cref{sec:results} presents controlled experimental results comparing binary and pass-rate rewards; \Cref{sec:analysis} provides mechanistic analyses of reward density and gradient directions; \Cref{sec:conclusion} concludes, with additional details in the appendix.

\section{Preliminaries}\label{sec:preliminary}
In this section, we formalize RL from unit-test feedback and introduce the policy-gradient methods and evaluation metrics used in experiments.

\subsection{Problem Formulation}
\label{subsec:formulation}
Let $\mathcal{X}$ denote the space of programming problems, and let $\mathcal{Y}$ denote the space of candidate programs.
There is a distribution $\mathcal{D}$ over the problem space $\mathcal{X}$ and a parametric policy $\pi_\theta(\cdot \mid x)$ implemented by an LLM, where $x \sim \mathcal{D}$.
Given a reward function $R(x,y)$, the training objective of RL can be formulated as:
\begin{equation}
  J(\theta) \triangleq \mathbb{E}_{x \sim \mathcal{D}} \mathbb{E}_{y \sim \pi_\theta(\cdot \mid x)} \bigl[ R(x,y) \bigr].
\end{equation}
Policy-gradient methods optimize $J(\theta)$ using rollout samples $(x,y)$ and their corresponding rewards $R(x,y)$. 

In RLVR, unit tests are always used to verify the correctness of generated programs.
For each problem $x$, there is a finite set of unit tests $\mathcal{T}_x = \{t_{x,1},\dots,t_{x,K_x}\}$.
Executing a program $y$ on a test $t_{x,k}$ produces a Boolean outcome $s_{x,k}(y) \in \{0,1\}$.
Therefore, the \emph{test-case pass rate} of code $y$ on problem $x$ can be written as:
\begin{equation}
  r(x,y) \triangleq \frac{1}{K_x} \sum_{k=1}^{K_x} s_{x,k}(y) \in \Bigl\{0,\frac{1}{K_x},\frac{2}{K_x},\dots,1\Bigr\}.
  \label{eq:pass-rate}
\end{equation}

Specifically, we consider the following reward choices for $R(x,y)$ in RL objective:

\noindent\textbf{Binary Reward.}
The conventional approach employs a sparse reward that distinguishes only perfect solutions, \emph{i.e.}, whether program $y$ passes all test cases of problem $x$:
\begin{equation}
  R_{\mathrm{bin}}(x,y) \triangleq \mathbb{I}\bigl[r(x,y)=1\bigr].
  \label{eq:binary-reward}
\end{equation}

\noindent\textbf{Pass-Rate Reward.}
To provide denser feedback, the pass-rate reward directly uses the test-case pass rate:
\begin{equation}
  R_{\mathrm{pass}}(x,y) \triangleq r(x,y).
  \label{eq:pass-rate-reward}
\end{equation}

\subsection{Policy-Gradient Methods}\label{subsec:pg-methods}
We employ two commonly used critic-free policy-gradient methods for RL fine-tuning: Group relative policy optimization~(GRPO) and reinforce leave-one-out~(RLOO).

\textbf{GRPO.}
GRPO~\citep{shao2024deepseekmath} is a group-wise policy-gradient method.
For a given problem $x$, the policy samples $N$ i.i.d.\ responses $y_1,\dots,y_N \sim \pi_\theta(\cdot \mid x)$ and computes corresponding rewards $r_i \triangleq R(x,y_i)$.
The policy is updated by maximizing a clipped surrogate objective:
\begin{equation}
  \mathcal{J}_{\mathrm{GRPO}}(\theta) = \mathbb{E}\Bigl[\min\bigl(\rho_i A_i,\; \mathrm{clip}_{1-\epsilon}^{1+\epsilon}(\rho_i)\, A_i\bigr)\Bigr],
  \label{eq:grpo-loss}
\end{equation}
where $\rho_i = \pi_\theta(y_i \mid x) / \pi_{\theta_{\mathrm{old}}}(y_i \mid x)$ is the importance ratio and $\epsilon$ is the clipping threshold.
The advantage $A_i$ is estimated by normalizing rewards within the group:
\begin{equation}
  A_i \triangleq \frac{r_i - \mu}{\sigma + \varepsilon},
  \label{eq:grpo-advantage}
\end{equation}
where $\mu$ and $\sigma$ denote the empirical mean and standard deviation of $\{r_1,\dots,r_N\}$, respectively, and $\varepsilon > 0$ is a small positive constant for numerical stability.

\textbf{RLOO.}
RLOO~\citep{ahmadian2024back} builds upon classical REINFORCE-style optimization by employing a leave-one-out baseline to obtain unbiased gradient estimates.
For a given problem $x$, the policy samples $N$ i.i.d.\ responses $y_1,\dots,y_N \sim \pi_\theta(\cdot \mid x)$ with rewards $r_i = R(x,y_i)$.
The policy is updated by maximizing:
\begin{equation}
  \mathcal{J}_{\mathrm{RLOO}}(\theta) = \mathbb{E}\bigl[A_i\bigr],
  \label{eq:rloo-loss}
\end{equation}
where $A_i$ is computed using a leave-one-out baseline:
\begin{equation}
  A_i \triangleq r_i - b_i, \quad b_i \triangleq \frac{1}{N-1} \sum_{j \neq i} r_j.
  \label{eq:rloo-advantage}
\end{equation}
Since $b_i$ depends only on $\{y_j\}_{j \neq i}$ and is independent of $y_i$, subtracting it does not introduce bias into the policy gradient estimator~\citep{sutton2018}.

\subsection{Evaluation Metric}\label{subsec:evaluation}
Following standard practice in code generation~\citep{lyu2025top,chen2021humaneval,kulal2019spoc}, we adopt \emph{pass@$k$} as the primary evaluation metric.
Given a problem $x$, pass@$k$ measures the probability that at least one of $k$ independently sampled programs passes all unit tests.
To obtain an unbiased estimate, we generate $n \geq k$ programs per problem and compute:
\begin{equation}
  \mathrm{pass@}k \triangleq 1 - \frac{\binom{n-c}{k}}{\binom{n}{k}},
  \label{eq:pass-at-k}
\end{equation}
where $c$ denotes the number of programs that pass all unit tests out of the $n$ generated programs.
This unbiased estimator is standard in code generation benchmarks, which avoids the systematic underestimation of the naive empirical formula $1-(1-\hat{p})^k$.

\section{Pass-Rate Reward Does Not Improve Code Generation Performance in RL}\label{sec:results}

\begin{table*}[ht]
\centering
\small
\caption{Test-set pass@$k$ (\%) comparing binary vs.\ pass-rate rewards across base models and algorithms. Below each pair, we report $\Delta$ (Pass-Rate$-$Binary), where \textcolor{red!65!black}{red} indicates Pass-Rate is higher and \textcolor{green!80!black}{green} indicates Binary is higher. DS-R1-7B denotes DeepSeek-R1-Distill-Qwen-7B; Qwen2.5-7B denotes Qwen2.5-7B-Instruct. We abbreviate LiveCodeBench as LCB and LeetCodeDataset as LC.}
\label{tab:main-table}
\begingroup
\setlength{\tabcolsep}{5pt}
\newcommand{\dpos}[1]{\textcolor{red!65!black}{#1}}
\newcommand{\dneg}[1]{\textcolor{green!80!black}{#1}}
\newcommand{\dzero}[1]{\textcolor{gray}{#1}}
\begin{tabular}{lllcccccccccccc}
\toprule
\multirow{2}{*}{\textbf{Model}} & \multirow{2}{*}{\textbf{Algorithm}} & \multirow{2}{*}{\textbf{Reward}} & \multicolumn{3}{c}{\textbf{Pass@1}} & \multicolumn{3}{c}{\textbf{Pass@4}} & \multicolumn{3}{c}{\textbf{Pass@8}} & \multicolumn{3}{c}{\textbf{Pass@16}} \\
\cmidrule(lr){4-6} \cmidrule(lr){7-9} \cmidrule(lr){10-12} \cmidrule(lr){13-15}
 & & & LCB & LC & Avg. & LCB & LC & Avg. & LCB & LC & Avg. & LCB & LC & Avg. \\
\midrule
\multirow[c]{3}{*}{DS-R1-7B} & \multirow{3}{*}{GRPO} & Binary & 33.5 & 46.1 & 40.6 & 42.1 & 57.1 & 50.6 & 45.5 & 61.1 & 54.3 & 48.6 & 64.5 & 57.6 \\
 &  & Pass-Rate & 33.8 & 46.4 & 40.9 & 41.0 & 57.0 & 50.0 & 43.8 & 60.3 & 53.1 & 46.3 & 62.7 & 55.6 \\
\cmidrule(lr){3-15}
 &  & \textit{$\Delta$ (PR$-$Bin)} & \dpos{+0.3} & \dpos{+0.3} & \dpos{+0.3} & \dneg{-1.1} & \dneg{-0.1} & \dneg{-0.6} & \dneg{-1.7} & \dneg{-0.8} & \dneg{-1.2} & \dneg{-2.3} & \dneg{-1.8} & \dneg{-2.0} \\
\midrule
\multirow{3}{*}{DS-R1-7B} & \multirow{3}{*}{RLOO} & Binary & 33.1 & 47.0 & 41.0 & 40.6 & 58.4 & 50.7 & 44.4 & 62.7 & 54.8 & 48.6 & 66.7 & 58.8 \\
 &  & Pass-Rate & 34.1 & 46.0 & 40.8 & 41.6 & 57.0 & 50.3 & 44.8 & 61.1 & 54.0 & 47.4 & 64.5 & 57.1 \\
\cmidrule(lr){3-15}
 &  & \textit{$\Delta$ (PR$-$Bin)} & \dpos{+1.0} & \dneg{-1.0} & \dneg{-0.2} & \dpos{+1.0} & \dneg{-1.4} & \dneg{-0.4} & \dpos{+0.4} & \dneg{-1.6} & \dneg{-0.8} & \dneg{-1.2} & \dneg{-2.2} & \dneg{-1.7} \\
\midrule
\multirow{3}{*}{Qwen3-4B} & \multirow{3}{*}{GRPO} & Binary & 37.7 & 53.2 & 46.4 & 44.2 & 61.4 & 53.9 & 46.7 & 64.3 & 56.6 & 49.1 & 66.7 & 59.1 \\
 &  & Pass-Rate & 36.5 & 50.2 & 44.2 & 43.2 & 59.4 & 52.4 & 45.3 & 62.3 & 54.9 & 46.9 & 64.5 & 56.8 \\
\cmidrule(lr){3-15}
 &  & \textit{$\Delta$ (PR$-$Bin)} & \dneg{-1.2} & \dneg{-3.0} & \dneg{-2.2} & \dneg{-1.0} & \dneg{-2.0} & \dneg{-1.5} & \dneg{-1.4} & \dneg{-2.0} & \dneg{-1.7} & \dneg{-2.2} & \dneg{-2.2} & \dneg{-2.3} \\
\midrule
\multirow{3}{*}{Qwen2.5-7B} & \multirow{3}{*}{GRPO} & Binary & 20.7 & 24.5 & 22.9 & 23.8 & 29.2 & 26.9 & 25.1 & 31.5 & 28.8 & 26.3 & 33.8 & 30.5 \\
 &  & Pass-Rate & 19.4 & 25.5 & 22.8 & 23.9 & 28.7 & 26.6 & 25.6 & 29.6 & 27.8 & 26.3 & 30.7 & 28.8 \\
\cmidrule(lr){3-15}
 &  & \textit{$\Delta$ (PR$-$Bin)} & \dneg{-1.3} & \dpos{+1.0} & \dneg{-0.1} & \dpos{+0.1} & \dneg{-0.5} & \dneg{-0.3} & \dpos{+0.5} & \dneg{-1.9} & \dneg{-1.0} & \dzero{0.0} & \dneg{-3.1} & \dneg{-1.7} \\
\bottomrule
\end{tabular}
\endgroup

\end{table*}


\begin{figure*}[t]
  \centering
  \begin{minipage}[t]{0.49\textwidth}
    \centering
    \includegraphics[width=\linewidth]{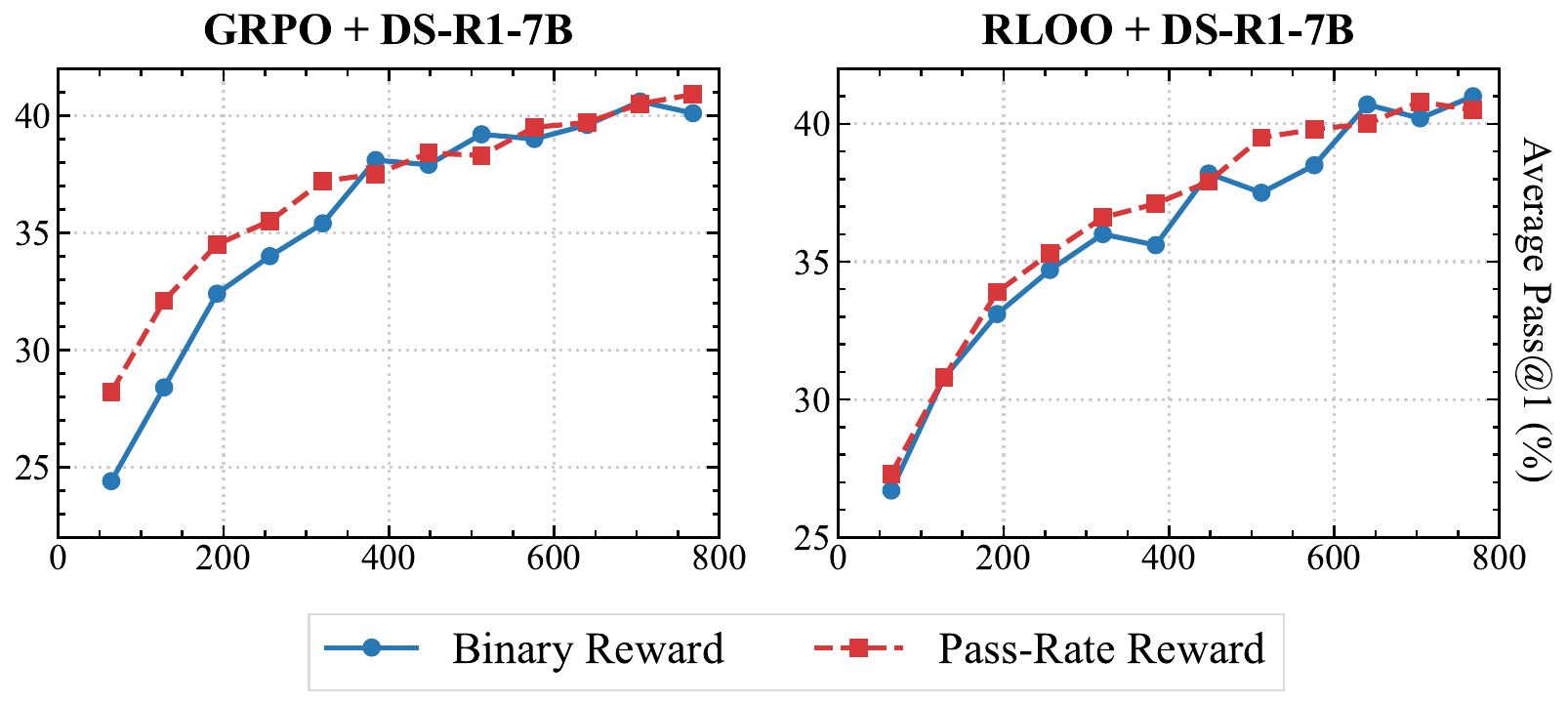}
  \end{minipage}
  \hfill
  \begin{minipage}[t]{0.49\textwidth}
    \centering
    \includegraphics[width=\linewidth]{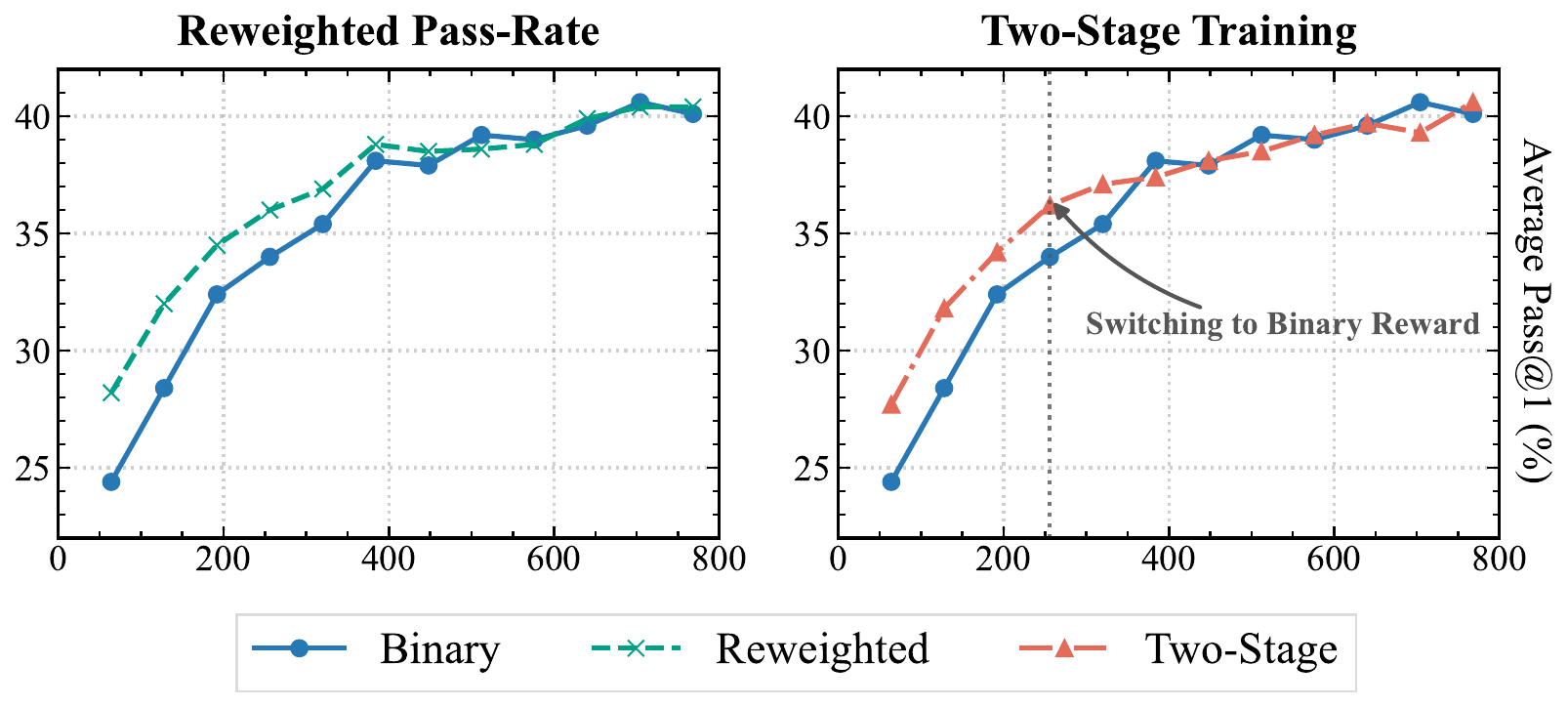}
  \end{minipage}
  \caption{Learning curves on DeepSeek-R1-Distill-Qwen-7B (pass@1 vs.\ training steps). Left: comparing binary and pass-rate rewards under GRPO and RLOO. Right: comparing pass-rate reward variants (reweighted pass-rate, two-stage) and binary reward under GRPO.}
  \label{fig:learning-curves}
\end{figure*}

We compare binary and pass-rate reward under rigorous controlled experimental settings.
Across different base models and RL algorithms, pass-rate reward does not reliably improve final performance over binary reward, and we find no consistent gains in pass@$k$ as shown in \Cref{tab:main-table}.
Below, we characterize these patterns across base models and algorithms, and examine whether pass-rate reward variants can consistently outperform the binary reward baseline.

\subsection{Experimental Setup}

\textbf{Datasets.} We train and evaluate on two code generation benchmarks: LiveCodeBench~\citep{livecodebench} and LeetCodeDataset~\citep{leetcodedataset}. LeetCodeDataset contains 2,641 training problems and 228 test problems, with test cases verified through sandbox execution against canonical solutions. LiveCodeBench contains 1,055 problems spanning May 2023 to April 2025. We use release v5 (problems up to January 4, 2025) as our training set, comprising 880 problems, and the incremental problems introduced in release v6 as our held-out test set, comprising 175 problems. We train on the union of both training splits and evaluate on the union of both test splits.

\textbf{Models.} We use DeepSeek-R1-Distill-Qwen-7B~\cite{guo2025deepseek-r1}, Qwen3-4B~\cite{yang2025qwen3} and Qwen2.5-7B-Instruct~\cite{yang2024qwen2} as base models for experiments.

\textbf{Algorithms.} We evaluate two representative RL algorithms, GRPO~\citep{zheng2025gspo} and RLOO~\citep{chen2025minimax}. While recent work on RL for LLMs has focused on refining importance-sampling corrections and
clipping mechanisms to stabilize off-policy updates~\citep{yu2025dapo,chen2025minimax,zheng2025gspo,gao2025soft}, we adopt a strict on-policy training regime to isolate the effect of reward design: each batch of rollouts is sampled from the current policy and consumed in a single gradient update, with $\theta_{\mathrm{old}}$ set to the policy that generated the batch so that the importance ratio is $\rho_i=1$ and the PPO-style clipping term is inactive in practice. Under this controlled setting, GRPO and RLOO share the same on-policy rollout sampling scheme and differ only in their advantage
estimators.

\textbf{Training Details.} We train for 768 gradient-update steps in all experiments. Training employs a batch size of 64 problems, with $N{=}16$ rollouts per problem.
The mini-batch size---i.e., the number of problems whose rollouts are consumed per gradient step---is also 64, ensuring strict on-policy training.
We set a constant learning rate of $1\mathrm{e}{-}6$ throughout training and apply neither KL regularization nor entropy control.
We use a temperature of 1.0 for both training and evaluation, following recent findings that a higher sampling temperature improves exploration and final performance in RL training~\cite{he2025skywork}.
To isolate the effect of reward shaping, all other hyperparameters are held constant across experiments. 

\subsection{Primary Results}

We first examine our primary configuration: DeepSeek-R1-Distill-Qwen-7B trained with GRPO\@.
As illustrated in \Cref{fig:learning-curves}, pass-rate reward shows a clear advantage over binary reward in the early stages of training.
However, this advantage diminishes as training progresses.
At convergence, pass-rate reward provides a slight improvement in pass@1, while underperforming the binary baseline on pass@$k$ for larger values of $k$ (\Cref{tab:main-table}).
Overall, pass-rate reward appears to accelerate early learning but does not raise the final performance ceiling.


\begin{table}[h]
\centering
\small
\caption{Task-level outcome overlap between binary and pass-rate rewards (GRPO + DeepSeek-R1-Distill-Qwen-7B) on the training set. Diagonal entries (bold) indicate agreement (97\%).}\label{tab:task-overlap}
\setlength{\tabcolsep}{4.8pt}
\begin{tabular}{c|cc|r}
\toprule
\diagbox{$R_\mathrm{Bin}$}{$R_\mathrm{pass}$} & Solved & Failed & \textit{Total}~(Agree\%)\\
\midrule
Solved & \cellcolor{gray!15}\textbf{2650} & 52 & 2702~(98\%) \\
Failed & 45 & \cellcolor{gray!15}\textbf{774} & 819~(95\%) \\
\midrule 
\textit{Total}~(Agree\%)& 2695~(98\%) & 826~(94\%) & 3521~(97\%) \\
\bottomrule
\end{tabular}
\end{table}

To assess whether the two rewards lead to different task-level capabilities beyond aggregate pass@$k$, we analyze per-task solvability on the training set for this primary setting.
We mark a task as \emph{solved} if at least one of sampled solutions passes all unit tests.
As shown in \Cref{tab:task-overlap}, the two runs agree on 97\% of tasks: 2,650 tasks are solved by both and 774 are failed by both.
Only 97 tasks differ between the two rewards, indicating that they learn highly similar sets of tasks and that pass-rate reward does not systematically unlock additional problems. The task-level overlap on the test set is reported in \Cref{appendix:task-level-solvability} and shows a similar pattern.

\subsection{Consistency Across Algorithms}
We employ two policy-gradient algorithms described in \cref{subsec:pg-methods}: GRPO and RLOO\@.
These two methods represent complementary bias--variance trade-offs in advantage estimation: GRPO's group-wise normalization yields stable gradients but may introduce bias, whereas RLOO's leave-one-out baseline provides unbiased estimates at the expense of higher variance. 
Evaluating both ensures our conclusions are not artifacts of a single estimator.

We replicate the experiments using RLOO\@. As shown in \Cref{fig:learning-curves}, pass-rate reward again converges to a level nearly identical to the binary baseline. At convergence, the binary reward achieves 41.0\% pass@1 and 58.8\% pass@16, while the pass-rate reward yields 40.8\% and 57.1\%, respectively (\Cref{tab:main-table}). This confirms that the lack of sustained improvement from pass-rate reward is not specific to a single reinforcement learning algorithm.

As a secondary observation, the early-stage acceleration seen under GRPO is muted under RLOO. A plausible explanation is that GRPO's variance normalization (\Cref{eq:grpo-advantage}) can amplify partial-pass advantages when full-pass samples are scarce, whereas RLOO lacks this scaling (\Cref{eq:rloo-advantage}). Overall, across GRPO and RLOO, pass-rate reward does not deliver durable gains, and early improvements are not robust to the choice of advantage estimator.

\subsection{Consistency Across Models}
We further investigate whether these findings generalize to alternative base models by training Qwen3-4B with GRPO. As shown in \Cref{tab:main-table}, pass-rate reward does not yield a performance gain over the binary baseline on Qwen3-4B; in fact, the binary reward achieves a higher pass@1 at convergence (46.4\% vs.\ 44.2\%).

To ensure our conclusions are not restricted to reasoning models, we extend our evaluation to Qwen2.5-7B-Instruct. We observe a consistent trend: the binary reward matches or marginally surpasses the pass-rate reward across all measured metrics as shown in \Cref{tab:main-table}. These results demonstrate that the absence of a meaningful performance gain from pass-rate rewards is a robust phenomenon that holds across both reasoning and non-reasoning models.

\subsection{Impact of Reward Variants}
Given that standard pass-rate reward does not reliably outperform binary reward, we investigate whether more sophisticated reward designs can bridge this gap. 
We evaluate two representative variants:

\textbf{Reweighted Pass-Rate Reward.}
Inspired by MiMo~\citep{xiaomi2025mimo}, we reweight test cases by difficulty to emphasize learning from more challenging cases:
\begin{equation}
  R_{\mathrm{reweight}}(x,y) = \frac{\sum_{k=1}^{K_x} w_k \cdot s_{x,k}(y)}{\sum_{k=1}^{K_x} w_k},
\end{equation}
where $w_k$ is the difficulty-based weight for test case $t_{x,k}$.
We estimate difficulty by measuring pass rates across models of varying capability and cluster test cases into Easy, Medium, and Hard categories. Details are provided in \Cref{appendix:test-case-diffculity}.

\textbf{Two-Stage Reward.}
When a rollout group contains no full-pass samples but includes partial-pass samples, pass-rate reward still provides non-zero gradients, facilitating early exploration while potentially biasing optimization away from full correctness.
To balance these trade-offs, we use pass-rate reward during an initial warmup phase and switch to binary reward thereafter.
Specifically, we train with pass-rate reward for the first 256 of 768 steps and switch to binary reward for the remaining training.

As illustrated in \Cref{fig:learning-curves}, both variants exhibit learning dynamics similar to standard pass-rate reward and fail to consistently surpass the binary baseline. For the DeepSeek-R1-Distill-Qwen-7B + GRPO configuration, reweighted pass-rate achieves 40.4\% pass@1 and 55.3\% pass@16, while two-stage training reaches 40.6\% pass@1 and 57.8\% pass@16. Neither approach offers a significant advantage over the pure binary reward baseline. These results suggest that the performance ceiling in RL for code generation is not dictated by reward density or weighting schemes, but rather by more fundamental factors, which we explore in the following section.

\section{Deep Analysis}\label{sec:analysis}

\Cref{sec:results} presents a counterintuitive finding: pass-rate reward and its variants do not improve code generation performance in RL over binary reward.
This raises a fundamental question: why doesn't pass-rate reward lead to better final performance while providing denser learning signals?

\subsection{Pass-Rate Reward Density Analysis}\label{sec:reward-density}

A natural hypothesis is that pass-rate reward may not provide a dense signal.
If test cases within a problem are highly correlated, generated solutions tend to pass or fail all tests together, causing the pass rates to collapse to a near-binary distribution.
We test this hypothesis by analyzing the distribution of pass-rate values observed during training for DeepSeek-R1-Distill-Qwen-7B with pass-rate reward.

\begin{figure}[t]
    \centering
    \includegraphics[width=\columnwidth]{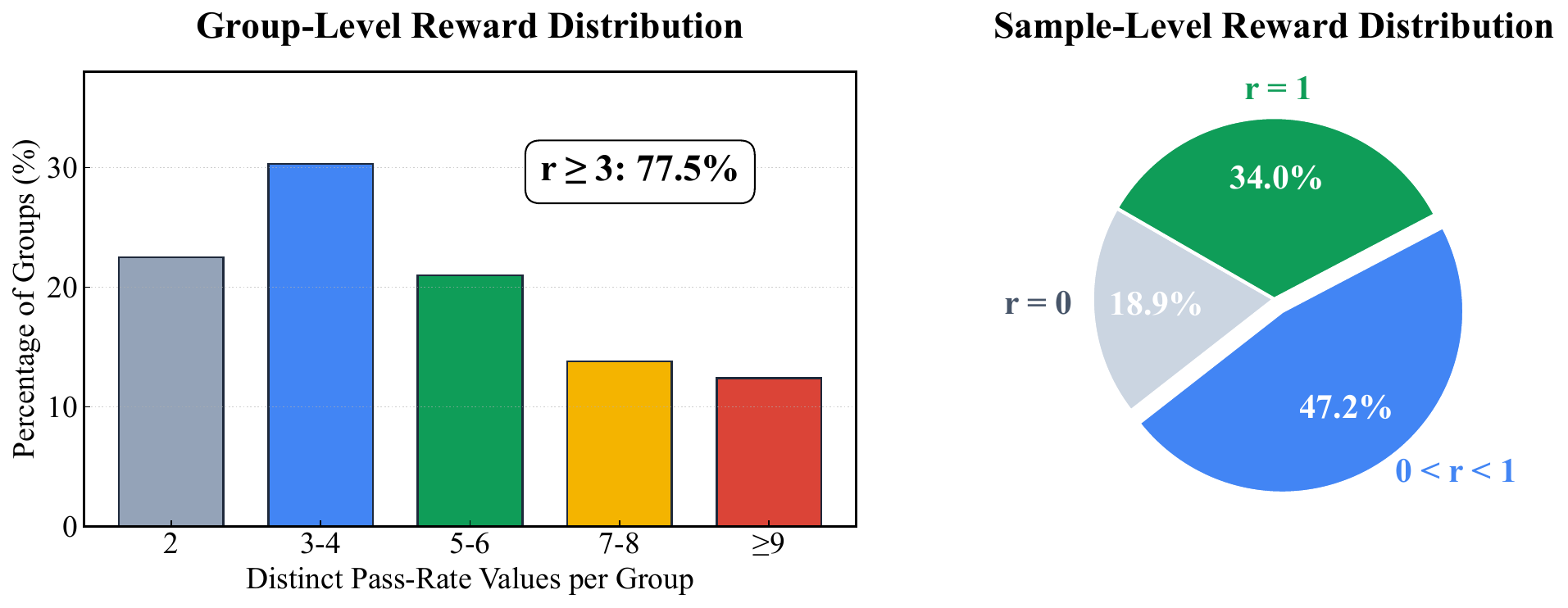}
    \caption{Pass-rate reward density analysis. Left: 77.5\% of groups contain 3+ distinct reward values. Right: Sample-level pass-rates span the full $[0,1]$ range, with 47.2\% intermediate values.}
    \label{fig:reward-density}
\end{figure}

Recall that GRPO samples a group of $N$ rollouts per task. Groups where all rollouts receive identical rewards produce zero gradients for both binary reward and pass-rate reward, so we exclude these ineffective groups and focus on effective groups where at least two distinct reward values exist.
As shown in the left panel of \Cref{fig:reward-density}, only 22.5\% of effective groups have two distinct reward values, while the remaining 77.5\% contain three or more distinct pass-rate values.
The right panel shows the sample-level reward distribution within effective groups, confirming substantial mass (47.2\%) on intermediate values.

\textbf{Pass-Rate Reward Provides Dense Feedback.}
These results rule out the ``insufficient density'' hypothesis in this setting: pass-rate rewards provide substantially denser feedback. Yet this denser signal does not translate into solving more tasks.
This motivates a deeper question: if the pass-rate reward is denser, why does it fail to improve final performance?
We argue that the answer lies in the \emph{gradient direction}, not reward density.

\subsection{Group-Level Gradient Analysis}\label{sec:gradient-direction}
The density analysis in \Cref{sec:reward-density} shows that pass-rate rewards provide abundant intermediate values, so one might expect them to provide a more informative signal when no full-pass sample is present.
Nevertheless, we observe no consistent improvement in final performance.
This indicates that reward density alone is insufficient: pass-rate rewards assign advantages to partial-pass samples, but those samples may or may not lie on a learning trajectory toward full-pass solutions. We therefore analyze the \emph{gradient direction} induced by pass-rate rewards.

Following the gradient probing methodology of \citet{ren2025learning}, we use a single-step probe update to characterize the local gradient direction induced by pass-rate rewards. This first-order probe captures the immediate effect of the policy-gradient update. While multi-step dynamics may accumulate non-linear effects, the probe provides a principled diagnostic for whether the pass-rate signal induces updates that locally point toward full correctness.

Formally, consider a code generation task $x$ and model parameters $\theta$.
Let $\{y_i\}_{i=1}^{N} \sim \pi_\theta(\cdot\!\mid\!x)$ denote a rollout group, and let $y_{\mathrm{full}}$ be a verified full-pass reference solution for the same task.
We say an update is \emph{toward full correctness} if it increases the likelihood of $y_{\mathrm{full}}$.
We estimate this via a within-task probe, measuring how updating $\theta$ changes the log-probability of $y_{\mathrm{full}}$.

\begin{figure}[t]
    \centering
    \IfFileExists{figures/group_level_delta_logprob_3way.pdf}{
        \includegraphics[width=\columnwidth]{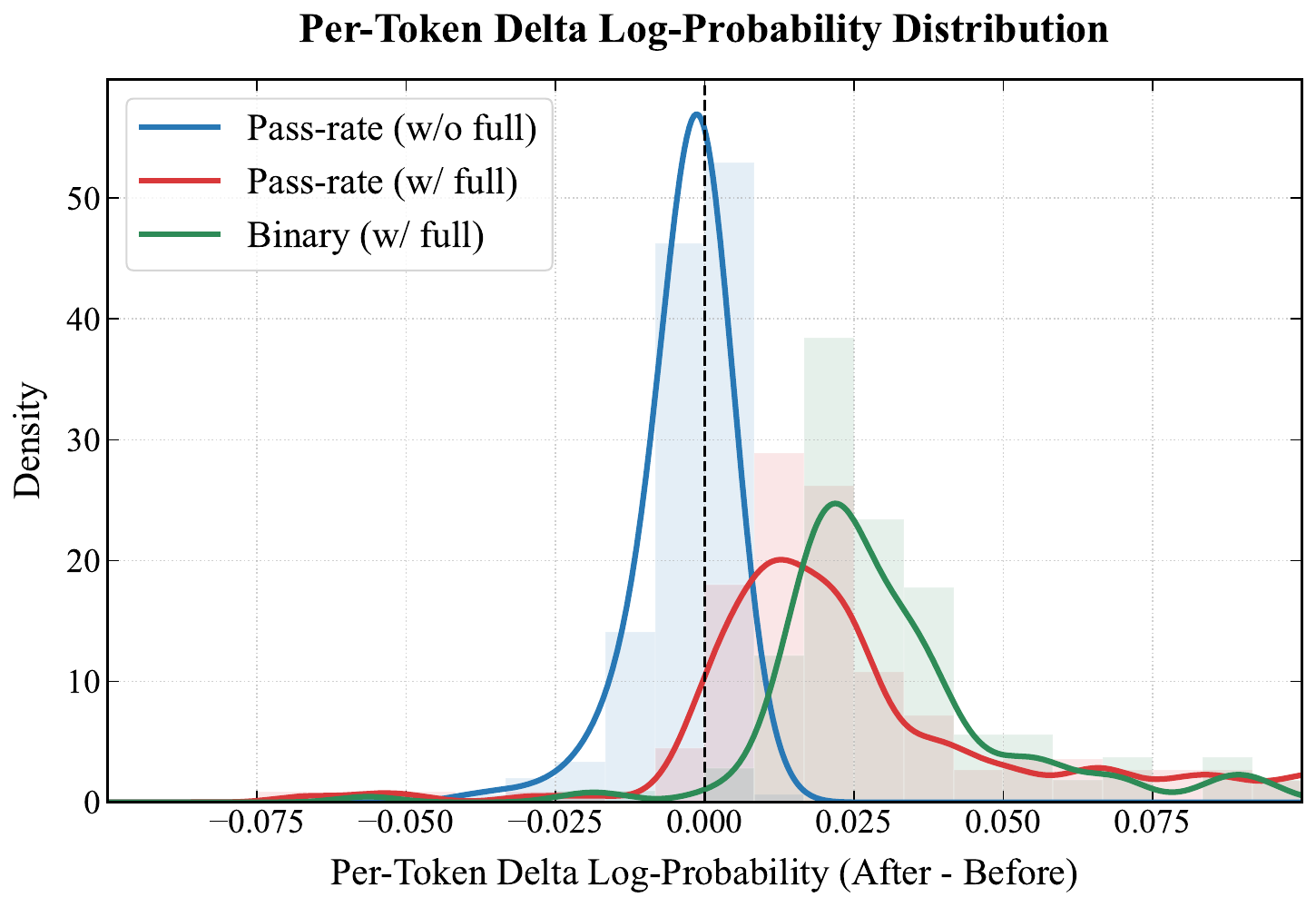}
    }{
        \fbox{\parbox[c][0.23\textheight][c]{\columnwidth}{\centering
	    \small Placeholder: group-level $\Delta \log p$ analysis \\ (to be added)}}
    }
    \caption{We plot the distribution of $\Delta_{\mathrm{grp}}$ (\Cref{eq:delta-grp}) induced by the probe update in \Cref{eq:group-update}. We report length-normalized $\Delta_{\mathrm{grp}}$ for visualization. The \emph{pass-rate, without-full} setting concentrates sharply near zero, indicating weak progress when only partial-pass samples are present, whereas augmenting the group with a full-pass reference solution (pass-rate or binary reward) yields a noticeably more positive shift.}\label{fig:delta-logprob-grp}
\end{figure}

To match the actual GRPO update, we apply a group-level probe update:
\begin{equation}\label{eq:group-update}
\theta' \leftarrow \theta + \eta \sum_{i=1}^N A_i\, \nabla_\theta \log \pi_\theta(y_i\!\mid\!x).
\end{equation}
Here $A_i$ is the GRPO advantage (\Cref{eq:grpo-advantage}), and $\eta>0$ is a small step size used only for probing.
\begin{equation}\label{eq:delta-grp}
\Delta_{\mathrm{grp}} \triangleq
\log \pi_{\theta'}(y_{\mathrm{full}}\!\mid\!x) - \log \pi_\theta(y_{\mathrm{full}}\!\mid\!x).
\end{equation}

Specifically, we run this probe on an early checkpoint from a GRPO + pass-rate reward training run.
We restrict to tasks whose rollout group contains no full-pass solution (the \emph{without-full} regime), which reflects the early-training setting where pass-rate reward is expected to provide the most benefit over binary reward.
For each task, we use a strong reasoning model (DeepSeek-R1~\citep{guo2025deepseek-r1}) to sample candidate reference solutions and keep the task only when we find a full-pass reference $y_{\mathrm{full}}$ that passes all test cases. This filtering yields 392 tasks for analysis.
Importantly, $y_{\mathrm{full}}$ is used only as a \emph{probe reference} to analyze gradient direction; it is never included in actual training.

To provide interpretable baselines on the same tasks and sampled groups, we compare these conditions:
\begin{enumerate}[leftmargin=*,label=(\alph*),nosep]
    \item \textbf{Pass-rate, without-full}: use only the sampled group with pass-rate rewards, which contains no full-pass reference solution in the group;
    \item \textbf{Pass-rate, with-full}: append a semantics-preserving rewritten variant $\tilde{y}_{\mathrm{full}}$ of $y_{\mathrm{full}}$ to the group and assign it reward $1$;
    \item \textbf{Binary, without-full}: omitted from \Cref{fig:delta-logprob-grp} because the resulting update is identically zero---under our zero-KL, zero-entropy setting, all samples receive reward $0$, yielding $A_i\!=\!0$ for all $i$ and thus $\Delta_{\mathrm{grp}}\!\equiv\!0$;
    \item \textbf{Binary, with-full}: append $\tilde{y}_{\mathrm{full}}$ to the group and assign it reward $1$ while other samples receive $0$.
\end{enumerate}
Conditions (b), (c), and (d) are counterfactual probes for analysis only; (c) is omitted from \Cref{fig:delta-logprob-grp} because the resulting update is identically zero, as noted above. We use the rewritten $\tilde{y}_{\mathrm{full}}$ to avoid evaluating $\Delta_{\mathrm{grp}}$ on a sequence that appears verbatim in the update group.
In all conditions, we compute the group-normalized advantages $A_i$ under the corresponding group, apply the group-level probe update in \Cref{eq:group-update}, and measure $\Delta_{\mathrm{grp}}$.

\textbf{Pass-Rate Reward Yields Negligible Progress Toward Full Correctness without Full-Pass Samples.}
\Cref{fig:delta-logprob-grp} isolates the setting that matters early in training: before any full-pass sample appears in a group.
Even though pass-rate provides non-binary rewards in this regime, the induced group-level update produces only a negligible change in the log-probability of the full-pass reference solution.
This explains why pass-rate alleviates reward sparsity but does not improve final performance: in the \emph{without-full} regime, the group-aggregated update yields near-zero changes in the log-probability of the full-pass reference, indicating that the overall optimization signal toward full correctness is weak and inconsistent.
In the next analysis, we zoom in to the sample level to diagnose how different samples within a group can induce opposing directional effects.

\subsection{Sample-Level Gradient Analysis}\label{sec:intra-group-conflict}

\Cref{fig:delta-logprob-grp} shows that when no full-pass sample is present, the group-level gradient update under pass-rate reward produces only a negligible transfer toward the full-pass reference solution. To understand why this transfer is weak, we probe the per-sample directional effects within the same rollout group.
For each $y_i$, we apply a single-sample probe update:
\begin{equation}
  \theta_i' \leftarrow \theta + \eta\, A_i\, \nabla_\theta \log \pi_\theta(y_i\!\mid\!x),
  \label{eq:single-sample-update}
\end{equation}
and measure the induced change in log-probability of the full-pass reference:
\begin{equation}
  \Delta_i \triangleq
  \log \pi_{\theta_i'}(y_{\mathrm{full}}\!\mid\!x) - \log \pi_\theta(y_{\mathrm{full}} \!\mid\!x).
  \label{eq:delta-logp}
\end{equation}
The sign of $\Delta_i$ indicates whether this update moves probability mass toward full correctness.

We focus on the practically relevant \emph{without-full} regime and study how different samples within the \emph{same} rollout group contribute to progress toward full correctness.
To illustrate this phenomenon, we randomly select 20 tasks from this regime and visualize the distribution of $\Delta_i$ values across the $N$ rollouts within each group.

\begin{figure}[t]
\centering
\includegraphics[width=\columnwidth]{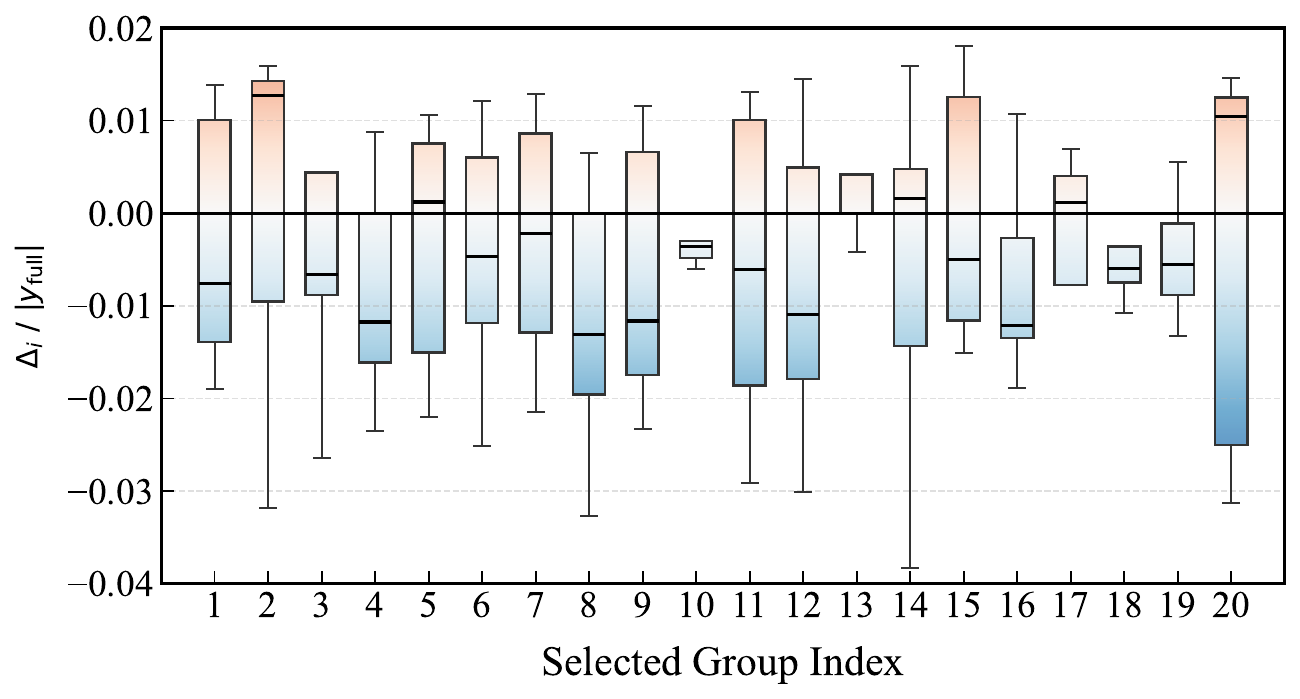}
\caption{Sample-level directional effects reveal intra-group gradient conflict. For 20 randomly selected tasks (each task corresponds to one rollout group in the without-full regime), we show boxplots of $\Delta_i$ (\Cref{eq:delta-logp}) over the $N$=16 samples in the group. We report length-normalized $\Delta_i$ for visualization. Many groups exhibit a mixed-sign distribution with both positive and negative $\Delta_i$, indicating ``push--pull'' updates within the same group.}
\label{fig:intra-group-conflict}
\end{figure}

\textbf{Intra-Group Gradient Conflict Leads to Weak Transfer Toward Full Correctness.}
\Cref{fig:intra-group-conflict} illustrates that within the same rollout group, different samples can induce updates in opposite directions: some increase the likelihood of a full-pass solution ($\Delta_i>0$), while others decrease it ($\Delta_i<0$).
When aggregated, these opposing effects destructively interfere, explaining the weak overall progress observed in \Cref{fig:delta-logprob-grp}.
We next quantify the prevalence of this conflict across tasks and diagnose its root cause.

\subsection{Root Cause of Intra-Group Gradient Conflict}\label{sec:root-cause}

\begin{figure}[t]
    \centering
    \includegraphics[width=\columnwidth]{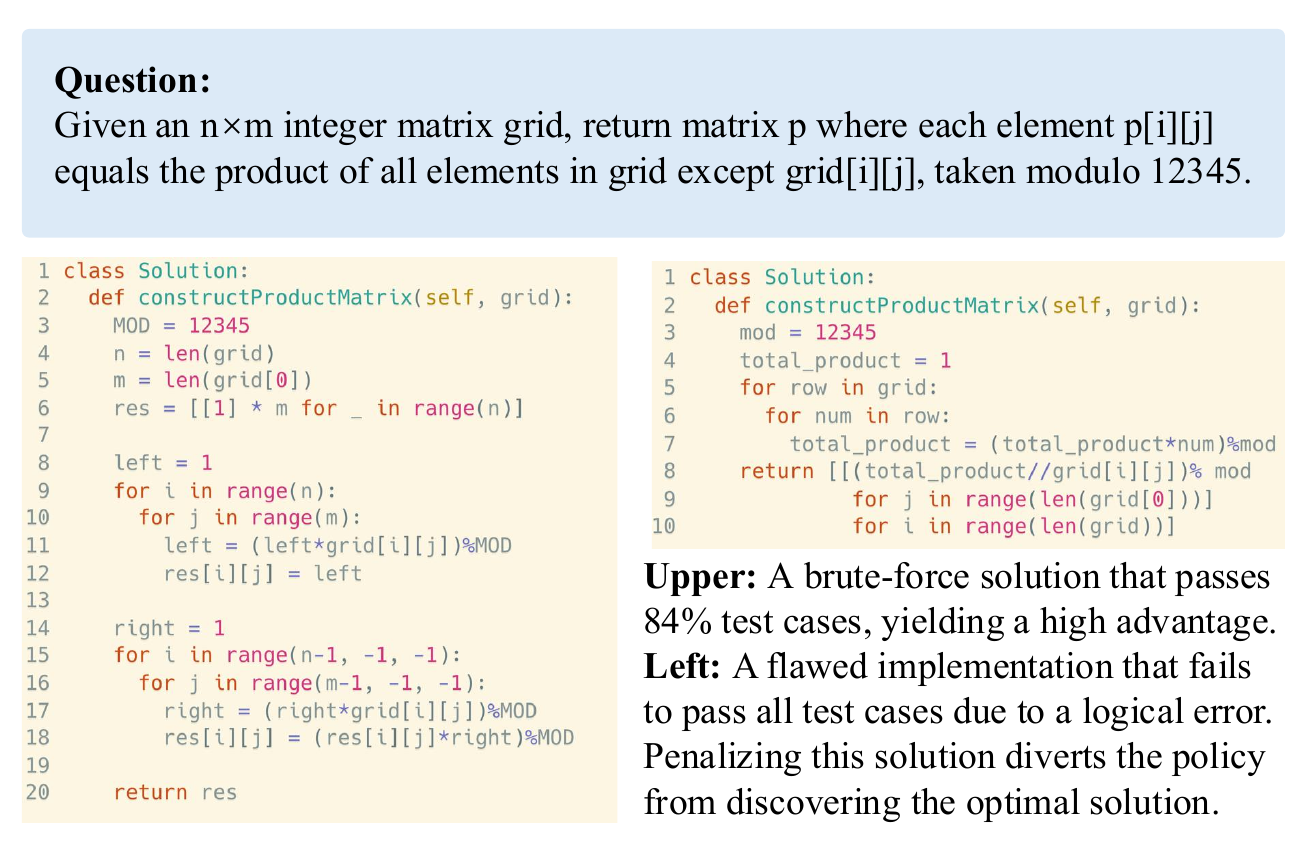}
    \caption{Case study of intra-group gradient conflict under pass-rate reward. For the same task, a high-pass but fundamentally wrong solution (upper right: it relies on large-integer division and passes 84\% of tests until timeout) can receive a larger advantage than a lower-pass but more helpful solution (left). As a result, the pass-rate update increases the likelihood of the harmful mode while decreasing the likelihood of the helpful mode, yielding a ``push--pull'' effect and weak overall progress toward full correctness.}
    \label{fig:case-study}
\end{figure}

To make the phenomenon concrete, we first provide a representative case study in \Cref{fig:case-study}.
It suggests that partial-pass samples can differ qualitatively in how they relate to a full-pass reference solution: some follow the reference approach and require only small implementation fixes, whereas others reflect qualitatively different and unproductive modes (e.g., a different core algorithm, heuristic overfitting to tests, or inefficiency leading to timeouts).

To quantify how often such mixed modes co-occur within the same rollout group, we run an LLM-as-judge analysis on the same 392 tasks used in our gradient probes in \Cref{sec:gradient-direction}.
For each task, we reuse the same sampled rollout group and full-pass reference solution $y_{\mathrm{full}}$ from the probe analysis.
We instruct DeepSeek-R1 to compare each incorrect sample $y_i$ against the full-pass reference solution and output a structured label indicating an error \emph{category} (e.g., \emph{implementation bug}, \emph{different core algorithm}, \emph{inefficient}, \emph{misread spec}, etc.); see \Cref{appendix:prompt_templates} for details.
For analysis, we group categories into \emph{helpful} versus \emph{harmful} (different core algorithm / heuristic or test-overfit / inefficient). Helpful samples correspond to solutions that are close to the reference and plausibly fixable with small patches, whereas harmful samples reflect qualitatively different modes.

Among the 392 tasks, 318 contain both helpful and harmful samples. Strikingly, in 116/318 tasks (36.5\%), the highest pass-rate harmful sample achieves a higher pass rate than the highest pass-rate helpful sample. This ``reversal'' shows that test-case pass rate is not a reliable proxy for progress toward ultimate correctness: even when helpful samples exist, pass rate can still prioritize algorithmically harmful samples.
We also quantify a pattern that directly reflects policy gradient updates. Specifically, we check whether a task contains (i) a harmful sample with positive advantage and (ii) a helpful sample with negative advantage. This is exactly the ``push--pull'' scenario where optimization increases the likelihood of a harmful mode while decreasing that of a helpful mode. We observe this advantage-sign conflict in 225/392 tasks (57.4\%), indicating that intra-group gradient conflict is common under pass-rate rewards.

\textbf{Test-Case Pass Rate Is an Unreliable Proxy for Proximity to Full Correctness.} Together with the gradient probes in \Cref{sec:intra-group-conflict}, these results provide a concrete mechanism for why pass-rate reward does not improve final performance: higher pass rate does not reliably indicate proximity to full correctness. Pass-rate reward can mis-rank qualitatively different failure modes, leading to intra-group gradient conflicts that hinder learning progress toward full correctness.

\section{Related Work}
	
\paragraph{Reinforcement Learning for LLMs.}
RL fine-tuning of LLMs typically employs policy gradient methods to align model behavior with reward signals from verifiers or human feedback.
Proximal Policy Optimization (PPO)~\cite{schulman2017ppo,ouyang2022training} follows the actor-critic paradigm: it learns a separate critic model to estimate state values, which serve as baselines for computing advantages via Generalized Advantage Estimation (GAE)~\cite{schulman2015gae}.
However, training and serving a dedicated critic network for LLMs is memory-intensive and complicates deployment.
To mitigate this overhead, recent work has developed critic-free alternatives that estimate baselines from sampled responses.
ReMax~\cite{li2023remax} approximates the baseline with the reward obtained by greedy decoding for each prompt.
Group Relative Policy Optimization (GRPO)~\cite{shao2024deepseekmath} estimates the baseline from the group average of multiple sampled responses and normalizes the centered rewards by the group standard deviation to stabilize training.
REINFORCE Leave-One-Out (RLOO)~\cite{ahmadian2024back} and Dr.GRPO~\cite{liu2025understanding} remove this standard deviation normalization and instead rely on leave-one-out or group-mean baselines, whereas REINFORCE++~\cite{hu2025reinforce++} adopts batch-mean and batch-standard-deviation normalization to improve stability and performance.
Recent approaches~\cite{yu2025dapo,zheng2025gspo,chen2025minimax,khatri2025scalerl} further explore improved clipping strategies and advantage estimation techniques.
Building on these critic-free policy-gradient methods (e.g., GRPO and RLOO), our work treats the RL algorithm as a controlled backbone and isolates an orthogonal axis: the choice between binary rewards and pass-rate rewards derived from unit tests.

\paragraph{Code Generation.}
Early LLMs such as GPT-3~\cite{brown2020gpt3} exhibited strong code generation capabilities even without specialized fine-tuning for coding tasks~\cite{chen2021humaneval,austin2021program}, revealing the potential of LLMs in this domain and inspiring extensive research focused on enhancing code generation performance.
Subsequent research explored various strategies, including high-quality dataset construction~\cite{luo2024wizardcoder,wei2024selfcodealign,liu2025revisiting}.
More recently, this field has shifted towards RL paradigms.
Early works like AlphaCode~\cite{li2022competition} demonstrated the effectiveness of RL combined with large-scale sampling and filtering.
Building on these ideas, recent work leverages execution feedback from unit tests to provide rewards and substantially improve code generation~\cite{gehring2025rlef,luo2025deepcoder}. 
Other methods~\cite{samadi2025scaling} combine RL with sophisticated sampling, clustering and reranking pipelines, enabling open-source LLMs to approach IOI~\cite{ioi2024} gold medal-level performance. 
RL has also been successfully scaled to complex software engineering tasks~\cite{jimenez2023swebench,deng2025swebenchpro}, where agents exploit feedback from full-repository execution environments~\cite{luo2025deepswe,jain2025r2egym} or patch similarity to ground-truth solutions~\cite{wei2025swerl}, yielding strong gains on multi-file bug fixing and issue resolution capabilities.

\paragraph{Pass-Rate Rewards.}
Pass-rate rewards provide a natural way to densify the sparse ``all-or-nothing'' signal used in RL for code generation, but their behavior remains relatively underexplored.
MiMo~\cite{xiaomi2025mimo} proposes a difficulty-driven code reward that groups test cases by difficulty and allocates partial credit across problem subtasks, thereby mitigating reward sparsity.
This design effectively turns the overall reward into a weighted sum of passed test cases, yet the work only briefly evaluates this component as part of a broader framework for post-training.
DELTA~\cite{sun2025deltas} uses pass-rate rewards as a warm-up stage followed by binary rewards on synthetic tasks to elicit ``algorithm unlocking'' behavior, rather than studying the effect of pass-rate rewards as the sole reward.
PRIME~\cite{cui2025process}, while primarily contributing an implicit process reward model trained from outcome labels, adopts the test-case pass rate (passed tests divided by total tests) as the outcome reward in its coding experiments and does not study pass-rate behavior in isolation.
Similarly, CURE~\cite{wang2025cure} employs pass-rate-based rewards in a co-evolutionary setting, where a single model jointly optimizes code generation and unit test generation, using pass rates primarily to score paired improvements in tests and solutions rather than to analyze the optimization properties of the reward itself.
In contrast, our work places pass-rate rewards at the center of the analysis and provides empirical and mechanistic evidence on why dense pass-rate reward does not optimize full correctness.

\section{Conclusion}\label{sec:conclusion}

In this work, we conducted controlled comparisons between the binary reward and the pass-rate reward across multiple base models, two critic-free RL algorithms (GRPO and RLOO), and several reward variants.
Our main finding is a consistent pattern within this regime: despite alleviating reward sparsity, the pass-rate reward does not improve the performance over binary reward.
To demystify this, we analyze the optimization signal induced by the pass-rate reward and reveal a core miscalibration: test-case pass rate is not a reliably monotonic proxy for correctness.

Overall, our results suggest that, in critic-free RL, simply densifying unit-test rewards via pass rate is insufficient to improve code generation.
Our findings provide a useful diagnostic lens and a strong baseline for future reward design in critic-free RL for code generation.
Future work should focus on reward designs and training signals that better align intermediate feedback with the ultimate objective of full correctness, while preserving the exploration benefits of dense supervision.

\section*{Impact Statement}

This paper presents work whose goal is to advance the field of machine learning by studying how pass-rate rewards affect reinforcement learning training for code generation.
There are many potential societal consequences of improving code generation systems, but none of them stand out in our work as requiring specific discussion beyond established considerations.

\bibliography{ref}
\bibliographystyle{icml2026}
\newpage
\appendix
\crefalias{section}{appendix}

\onecolumn

\section{Test-Case Reweighted Pass-Rate Reward}
\label{appendix:test-case-diffculity}

For the reweighted pass-rate reward (used in \Cref{sec:results}), we assign larger weights to harder test cases so that passing them contributes more to the overall reward.
Concretely, we first estimate a per-test-case difficulty signal from model pass rates, then map difficulty to a weight $w_k$ in the reweighted pass-rate reward:
\begin{equation}
  R_{\mathrm{reweight}}(x,y) = \frac{\sum_{k=1}^{K_x} w_k \cdot s_{x,k}(y)}{\sum_{k=1}^{K_x} w_k}.
\end{equation}

\paragraph{Heuristic difficulty labeling.}
We quantify test-case difficulty using pass rates computed from solutions generated by three models of varying capabilities: DeepSeek-R1~\cite{guo2025deepseek-r1}, QwQ~\cite{qwq32b}, and DeepSeek-R1-Distill-Qwen-7B~\cite{guo2025deepseek-r1}.
Algorithm~\ref{alg:tc-difficulty} summarizes the procedure.
For each task, we compute a weighted pass signal $p[j]$ for each test case $j$ by aggregating per-model test-case pass rates, using fixed model weights derived from each model's overall pass@1 on a calibration set.
We then label test cases as \textsc{Easy}/\textsc{Medium}/\textsc{Hard} based on within-task percentiles (33rd/67th), with two degenerate shortcuts: if all test cases are nearly always passed (resp.\ nearly always failed), we label all of them as \textsc{Easy} (resp.\ \textsc{Hard}).
In the reweighted reward, we assign higher weights to \textsc{Hard} cases and lower weights to \textsc{Easy} cases.

\begin{algorithm}[htbp]
\caption{Heuristic Test-Case Difficulty Labeling for Reweighting}
\label{alg:tc-difficulty}
\begin{algorithmic}[1]
\REQUIRE Task $x$, test-case index set $\mathcal{J}_x$, $\epsilon=0.1$.
\ENSURE Difficulty labels $\texttt{difficulty}[j]\in\{\textsc{Easy},\textsc{Medium},\textsc{Hard}\}$ for all $j\in\mathcal{J}_x$.

\STATE Generate multiple solutions for task $x$ using three models $m_1,m_2,m_3$.
\STATE Compute fixed model weights using pass@1: $w_i \leftarrow \mathrm{pass@1}(m_i)$ for $i\in\{1,2,3\}$ \COMMENT{computed on a calibration set}

\FORALL{$j \in \mathcal{J}_x$} 
    \STATE Compute per-model pass rates on test case $j$: $r_1[j], r_2[j], r_3[j]$ \COMMENT{aggregated over rollouts}
    \STATE $p[j] \leftarrow \dfrac{w_1 r_1[j] + w_2 r_2[j] + w_3 r_3[j]}{w_1+w_2+w_3}$ \COMMENT{weighted pass signal}
\ENDFOR

\IF{$\forall j\in\mathcal{J}_x:\; p[j] \ge 1-\epsilon$}
    \FORALL{$j\in\mathcal{J}_x$}
        \STATE $\texttt{difficulty}[j] \leftarrow \textsc{Easy}$ \COMMENT{nearly always passed}
    \ENDFOR
    \STATE \textbf{Return} $\texttt{difficulty}$
\ENDIF

\IF{$\forall j\in\mathcal{J}_x:\; p[j] \le \epsilon$}
    \FORALL{$j\in\mathcal{J}_x$}
        \STATE $\texttt{difficulty}[j] \leftarrow \textsc{Hard}$ \COMMENT{nearly always failed}
    \ENDFOR
    \STATE \textbf{Return} $\texttt{difficulty}$
\ENDIF

\STATE $\texttt{vals} \leftarrow \mathrm{sort}(\{p[j] : j\in\mathcal{J}_x\})$ \COMMENT{within-task sorting}
\STATE $q_{33} \leftarrow \mathrm{percentile}(\texttt{vals}, 33)$; \ \ $q_{67} \leftarrow \mathrm{percentile}(\texttt{vals}, 67)$ \COMMENT{within-task quantiles}

\FORALL{$j \in \mathcal{J}_x$}
    \IF{$p[j] \ge q_{67}$}
        \STATE $\texttt{difficulty}[j] \leftarrow \textsc{Easy}$
    \ELSIF{$p[j] < q_{33}$}
        \STATE $\texttt{difficulty}[j] \leftarrow \textsc{Hard}$
    \ELSE
        \STATE $\texttt{difficulty}[j] \leftarrow \textsc{Medium}$
    \ENDIF
\ENDFOR

\STATE \textbf{Return} $\texttt{difficulty}$
\end{algorithmic}
\end{algorithm}

\section{Implementation Details}\label{appendix:implementation-details}
We implement our RL training pipeline upon the verl framework~\cite{sheng2025hybridflow}, which provides efficient rollout sampling, reward computation, and distributed training capabilities.


\textbf{Evaluation Protocols.}
We evaluate checkpoints using the sampling temperature $T=1.0$ with nucleus sampling ($p=0.95$). We generate $N=16$ samples per problem to estimate pass@$k$, following prior work~\cite{liu2025coder1}.

\textbf{Code Execution Environment.}
We execute generated code in a sandboxed environment using \texttt{firejail}, with resource limits enforced to prevent infinite loops and excessive memory consumption.
For each problem, we run all associated test cases and compute rewards based on execution outcomes.
Following standard practice in competitive programming~\citep{domjudge,icpcguidelines}, we set per-problem timeout thresholds using an adaptive scheme based on the canonical solution's runtime:
\begin{equation}
  t_{\mathrm{timeout}} =
    \mathrm{clamp}(S \cdot t_{\mathrm{canon}},\, T_{\min},\, T_{\max})
\end{equation}
where $t_{\mathrm{canon}}$ is the canonical solution's execution time, $S{=}3$ is a time limit multiplier, $T_{\min}{=}10\mathrm{s}$, and $T_{\max}{=}30\mathrm{s}$.
For problems without a canonical solution, we default to $T_{\min}$.
This prevents both premature termination of correct but slow solutions and excessive waiting on pathological inputs.

\section{Task-Level Solvability}\label{appendix:task-level-solvability}

To complement the aggregate pass@$k$ results, we report task-level solvability overlap on the held-out test set for our primary setting (GRPO + DeepSeek-R1-Distill-Qwen-7B).
For each task, we mark it as \emph{solved} if at least one of $N{=}16$ sampled solutions passes all unit tests, and \emph{failed} otherwise.
As shown in \Cref{tab:task-overlap-test}, the binary reward and the pass-rate reward agree on 94\% of test tasks (216 solved by both and 164 failed by both), indicating that they largely succeed and fail on the same problems.
Among the 23 disagreement cases, binary solves slightly more tasks overall (231 vs.\ 224), with 15 tasks solved only by binary and 8 tasks solved only by pass-rate.
\Cref{tab:cross_benchmark} further breaks down these outcomes by benchmark and shows a similar pattern on both LeetCodeDataset and LiveCodeBench.

\begin{table}[h]
\centering
\begin{subtable}[t]{0.48\linewidth}
\centering
\small
\caption{Test-set outcome overlap between binary and pass-rate rewards (GRPO + DeepSeek-R1-Distill-Qwen-7B). Diagonal entries (bold) indicate agreement (94\%).}
\label{tab:task-overlap-test}
\begingroup
\renewcommand{\arraystretch}{1.3}
\setlength{\tabcolsep}{6pt}
\begin{tabular}{llrrr}
\toprule
& & \multicolumn{2}{c}{\textbf{Pass-Rate R.}} & \\
\cmidrule(lr){3-4}
& & Solved & Failed & \textit{Total} \\
\midrule
\multirow{2}{*}{\textbf{Binary R.}} & Solved & \cellcolor{gray!15}\textbf{216} & 15 & 231 \\
& Failed & 8 & \cellcolor{gray!15}\textbf{164} & 172 \\
\midrule 
& \textit{Total} & 224 & 179 & 403 \\
\bottomrule
\end{tabular}
\endgroup
\end{subtable}
\hfill
\begin{subtable}[t]{0.48\linewidth}
\centering
\small
\caption{Per-benchmark confusion percentages for binary vs pass-rate rewards.}
\label{tab:cross_benchmark}
\begingroup
\renewcommand{\arraystretch}{1.2}
\setlength{\tabcolsep}{8pt}
\begin{tabular}{cc|cc} 
\toprule
\textbf{Binary} & \textbf{Pass-Rate} & \textbf{LC} & \textbf{LCB} \\
\midrule
\cmark & \cmark & 61.0\% & 44.0\% \\
\cmark & \xmark & 4.4\% & 2.9\% \\
\xmark & \cmark & 1.8\% & 2.3\% \\
\xmark & \xmark & 32.9\% & 50.9\% \\
\bottomrule
\end{tabular}
\endgroup
\end{subtable}
\end{table}

\section{Reproducibility}
\label{appendix:reproducibility}

\textbf{Data.}
All datasets used in this paper are publicly available. We use the official releases of LiveCodeBench~\citep{livecodebench} (training: release v5, test: release v6) and LeetCodeDataset~\citep{leetcodedataset} (temporal split with August 4, 2024 as the cutoff), as described in \Cref{sec:results}.

\textbf{Code.}
Our full training and evaluation pipeline, including reward implementations (binary, pass-rate, and variants), execution-based reward computation, and scripts to reproduce our experiments, is provided in the supplementary material.

\section{Robustness of the LLM-as-Judge Analysis}\label{appendix:llm-judge-robustness}

The helpful/harmful categorization in \Cref{sec:root-cause} is produced by an LLM-as-judge that compares each incorrect rollout against a full-pass reference solution. To verify that the reversal and intra-group gradient conflict signals are not artifacts of a particular prompt phrasing or judge model, we conduct a prompt and model sensitivity sweep over four prompt variants and two judge models, yielding eight $(\text{prompt}, \text{model})$ configurations.

\textbf{Setup.} The four prompt variants are:
\begin{itemize}[leftmargin=*,nosep,topsep=0pt]
    \item \texttt{baseline}: the prompt used in \Cref{sec:root-cause} (see \Cref{appendix:prompt_templates}).
    \item \texttt{no\_pass\_rate}: removes the candidate's pass rate from the prompt context to reduce judge bias from the candidate's apparent score.
    \item \texttt{strict\_rubric}: tightens the rubric to label a candidate as harmful only when there is strong evidence of qualitatively different failure modes.
    \item \texttt{skeptical\_near\_miss}: makes the rubric more skeptical about labeling a candidate as a near-miss bug when the candidate's implementation differs nontrivially from the reference.
\end{itemize}
The two judge models are DeepSeek-R1~\cite{guo2025deepseek-r1} and GPT-5.4. Following \Cref{sec:root-cause}, we report two task-level metrics: the \emph{reversal rate}, defined as the fraction of with-both tasks (those containing at least one helpful and one harmful sample) in which the highest-pass-rate harmful sample exceeds the highest-pass-rate helpful sample; and the \emph{conflict rate}, defined as the fraction of any-judged tasks that contain both a harmful sample with positive advantage and a helpful sample with negative advantage. The robustness sweep covers the same 392 tasks as the paper baseline in \Cref{sec:root-cause}.

\textbf{Results.} \Cref{tab:llm-judge-robustness} reports both metrics across all configurations. The reversal signal stays in the $28.8\%$--$37.4\%$ range across every prompt variant and judge model, comparable to the paper baseline of $36.5\%$. The conflict signal becomes more conservative under stricter rubrics and under GPT-5.4, but remains clearly positive (between $21.5\%$ and $57.5\%$) in every configuration. We conclude that neither the reversal nor the intra-group gradient conflict signal depends on the specific judge model or prompt phrasing.

\begin{table}[h]
\centering
\small
\caption{Robustness of the LLM-as-judge analysis across four prompt variants and two judge models. Reversal and conflict are defined as in \Cref{sec:root-cause}. The paper baseline (top row) corresponds to the configuration used in the main paper.}
\label{tab:llm-judge-robustness}
\begin{tabular}{llrrrrrr}
\toprule
Judge Model & Prompt & Any Judged & With Both & Reversal & Reversal Rate & Conflict & Conflict Rate \\
\midrule
\multicolumn{2}{l}{\emph{Paper baseline (DeepSeek-R1, baseline prompt)}} & 392 & 318 & 116 & $36.5\%$ & 225 & $57.5\%$ \\
\midrule
DeepSeek-R1 & \texttt{no\_pass\_rate}        & 392 & 270 & 101 & $37.4\%$ & 164 & $41.9\%$ \\
DeepSeek-R1 & \texttt{strict\_rubric}        & 392 & 211 &  68 & $32.2\%$ &  85 & $21.7\%$ \\
DeepSeek-R1 & \texttt{skeptical\_near\_miss} & 392 & 251 &  92 & $36.7\%$ & 152 & $38.9\%$ \\
\midrule
GPT-5.4 & \texttt{baseline}              & 392 & 212 &  61 & $28.8\%$ &  90 & $23.0\%$ \\
GPT-5.4 & \texttt{no\_pass\_rate}        & 392 & 215 &  74 & $34.4\%$ & 104 & $26.6\%$ \\
GPT-5.4 & \texttt{strict\_rubric}        & 392 & 200 &  72 & $36.0\%$ &  84 & $21.5\%$ \\
GPT-5.4 & \texttt{skeptical\_near\_miss} & 392 & 195 &  73 & $37.4\%$ &  93 & $23.8\%$ \\
\bottomrule
\end{tabular}
\end{table}

\section{Limitations}
Our study has several limitations that warrant further investigation.
First, our \emph{reweighted pass-rate reward} is a best-effort approximation inspired by MiMo's difficulty-weighted code reward~\cite{xiaomi2025mimo}. Since key implementation details of MiMo are not publicly available, our heuristic reweighting may not faithfully replicate MiMo's reward. 
Second, we intentionally keep training hyperparameters fixed across experiments to isolate the effect of reward choices. As a result, we do not exhaustively tune hyperparameters for each reward variant individually.
Finally, our experiments focus on Python programming tasks from competitive programming benchmarks. On other settings such as agentic software development~\cite{jimenez2023swebench,luo2025deepswe}, the dynamics of reward choices remain to be explored.

\section{Prompt Templates}\label{appendix:prompt_templates}

We provide the prompt templates used in our experiments in \Cref{fig:prompt-template-starter-code,fig:prompt-template-without-starter-code,fig:prompt-template-error-analysis}. 
The first two templates are used for training and evaluation in our RL experiments, with and without starter code, respectively, while the third template is used for error analysis of incorrect solutions in \Cref{sec:root-cause}.

\definecolor{academicblue}{RGB}{60, 100, 180} 
\definecolor{lightblue}{RGB}{240, 245, 250}

\begin{figure}[h]
\centering
\begin{tcolorbox}[
    colback=lightblue,
    colframe=academicblue,
    coltitle=white,
    fonttitle=\bfseries\sffamily\small,
    boxrule=0.5pt,
    arc=2pt,
    left=8pt, right=8pt, top=6pt, bottom=6pt,
    title=Prompt Template with Starter Code
]

\textbf{System:} \\
\texttt{You are a helpful programming assistant. The user will ask you a question and you 
as the assistant solve it. The assistant first thinks how to solve the task through 
reasoning and then provides the user with the final answer. The reasoning process 
and answer are enclosed within <think>...</think> and <answer>...</answer> tags, 
respectively.}

\vspace{1em}
\textbf{User Input:} \\
\texttt{Please solve the programming task below using a self-contained code snippet in a markdown code block. \\ \\
\#\#\# Question:\\}
\{question\} \\\\
\texttt{\#\#\# Format: Read the inputs from stdin solve the problem and write the answer to stdout (do not directly test on the sample inputs). Enclose your code within delimiters as follows. Ensure that when the python program runs, it reads the inputs, runs the algorithm and writes output to STDOUT..\\
'''python\\}
\{starter\_code\} \\
\texttt{'''} \\\\
\texttt{\#\#\# Answer: (use the provided format with backticks)}

\end{tcolorbox}
\caption{Prompt template with starter code used in our training and evaluation.}
\label{fig:prompt-template-starter-code}
\end{figure}

\begin{figure}[h]
\centering
\begin{tcolorbox}[
    colback=lightblue,
    colframe=academicblue,
    coltitle=white,
    fonttitle=\bfseries\sffamily\small,
    boxrule=0.5pt,
    arc=2pt,
    left=8pt, right=8pt, top=6pt, bottom=6pt,
    title=Prompt Template without Starter Code
]

\textbf{System:} \\
\texttt{You are a helpful programming assistant. The user will ask you a question and you 
as the assistant solve it. The assistant first thinks how to solve the task through 
reasoning and then provides the user with the final answer. The reasoning process 
and answer are enclosed within <think>...</think> and <answer>...</answer> tags, 
respectively.}

\vspace{1em}
\textbf{User Input:} \\
\texttt{Please solve the programming task below using a self-contained code snippet in a markdown code block. \\ \\
\#\#\# Question:\\}
\{question\} \\\\
\texttt{\#\#\# Format: You will use the following starter code to write the solution to the problem and enclose your code within delimiters.\\
'''python\\ \# YOUR CODE HERE\\}
\texttt{'''} \\\\
\texttt{\#\#\# Answer: (use the provided format with backticks)}

\end{tcolorbox}
\caption{Prompt template without starter code used in our training and evaluation.}
\label{fig:prompt-template-without-starter-code}
\end{figure}

\begin{figure}[h]
\centering
\begin{tcolorbox}[
    colback=lightblue,
    colframe=academicblue,
    coltitle=white,
    fonttitle=\bfseries\sffamily\small,
    boxrule=0.5pt,
    arc=2pt,
    left=8pt, right=8pt, top=6pt, bottom=6pt,
    title=Prompt Template for Error Analysis
]

\textbf{System:} \\
\texttt{You are an expert code reviewer doing *error analysis*. \\
You are given: \\
- a programming PROBLEM \\
- an ORACLE solution (assume it is fully correct and passes all tests) \\
- a CANDIDATE solution (known to be wrong: it fails at least one hidden test\\\\
Your task is NOT to solve the problem. Instead, compare ORACLE vs CANDIDATE and classify WHY the candidate is wrong, relative to the oracle.\\\\
Return JSON only (no markdown) with keys:\\
- category: one of ["near\_miss\_bug","different\_core\_algorithm",\\
"heuristic\_or\_test\_overfit","inefficient\_or\_tle","misread\_spec",\\"incomplete\_solution","format\_or\_runtime","unclear"]\\
- same\_core\_algorithm: boolean (whether candidate uses essentially the same core algorithm as oracle)\\
- fix\_effort: one of ["small\_patch","moderate\_changes","rewrite"]\\
- evidence: 2-4 short lines pointing to concrete differences (no new code)\\
- confidence: float in [0,1]\\\\
Guidance:\\
- Prefer "near\_miss\_bug" when the candidate is basically the oracle approach but has a logic/edge-case/implementation bug.\\
- Prefer "different\_core\_algorithm" when the main approach differs (e.g. greedy vs DP, wrong model, wrong invariant).\\
- Prefer "heuristic\_or\_test\_overfit" when the candidate looks like a heuristic / pattern-matching / special-casing that could pass many tests but is not generally correct.\\
- Prefer "inefficient\_or\_tle" when the approach is plausibly correct but asymptotically too slow vs expected constraints.\\
- Prefer "misread\_spec" when the candidate solves a different problem than asked.\\
- Prefer "format\_or\_runtime" only for obvious runtime/format issues.}

\vspace{1em}
\textbf{User Input:} \\
\texttt{PROBLEM:\\}
\{problem\} \\\\
\texttt{ORACLE\_SOLUTION:\\}
\{oracle\_solution\} \\\\
\texttt{CANDIDATE\_SOLUTION:\\}
\{candidate\_solution\} \\\\
\texttt{Candidate test-case pass rate (context only):\\}
\{candidate\_pass\_rate\} \\\\

\end{tcolorbox}
\caption{Prompt template for error analysis.}
\label{fig:prompt-template-error-analysis}
\end{figure}


\end{document}